\pdfoutput=1
\documentclass[11pt]{article}

\usepackage[]{ACL2023}

\usepackage{times}
\usepackage{latexsym}

\usepackage[T1]{fontenc}
\usepackage[utf8]{inputenc}

\usepackage{microtype}

\usepackage{inconsolata}

\usepackage{graphicx}
\usepackage{fancyvrb}

\usepackage{tabularx}
\usepackage{multirow}
\usepackage{tablefootnote}
\usepackage{booktabs}
\usepackage{amssymb}
\usepackage{amsmath}
\usepackage{hyperref}
\usepackage{listings}
\usepackage{xcolor}
\usepackage{sourcecodepro}
\usepackage[T1]{fontenc}
\usepackage{logicproof}
\usepackage{pifont}% http://ctan.org/pkg/pifont
\usepackage{dblfloatfix}
\newcommand{\xmark}{\ding{55}}%

\definecolor{codegreen}{rgb}{0,0.6,0}
\definecolor{codegray}{rgb}{0.5,0.5,0.5}

\definecolor{backcolour}{RGB}{245,248,250}
\definecolor{emph}{RGB}{166,88,53}
\definecolor{nightblue}{RGB}{9,49,105}
\definecolor{keywords}{RGB}{207,33,46}
\definecolor{lightpurple}{RGB}{130,81,223}

\lstdefinestyle{mystyle}{
    backgroundcolor=\color{backcolour},   
    commentstyle=\color{codegreen},
    keywordstyle=\color{keywords},
    stringstyle=\color{nightblue},
    basicstyle=\fontsize{7}{8}\ttfamily,
    breakatwhitespace=true,         
    breaklines=true,                 
    captionpos=b,                    
    keepspaces=true,                 
    numberstyle=\tiny\color{codegray},
    numbersep=2pt,                  
    showspaces=false,                
    showstringspaces=false,
    showtabs=false,                  
    tabsize=2,
    emph={dspy},
    emphstyle={\color{lightpurple}},
    linewidth=1\columnwidth,
    frame=tb,    
    xrightmargin=0pt,
    xleftmargin=0.23cm,
    aboveskip=0.2cm,
    belowskip=0.1cm,
}

\colorlet{punct}{red!60!black}
\definecolor{delim}{RGB}{20,105,176}
\colorlet{numb}{magenta!60!black}

\lstdefinelanguage{json}{
    basicstyle=\fontsize{7}{8}\ttfamily,
    stepnumber=1,
    numbersep=8pt,
    showstringspaces=false,
    breaklines=true,
    frame=lines,
    backgroundcolor=\color{backcolour},   
    literate=
     *{0}{{{\color{numb}0}}}{1}
      {1}{{{\color{numb}1}}}{1}
      {2}{{{\color{numb}2}}}{1}
      {3}{{{\color{numb}3}}}{1}
      {4}{{{\color{numb}4}}}{1}
      {5}{{{\color{numb}5}}}{1}
      {6}{{{\color{numb}6}}}{1}
      {7}{{{\color{numb}7}}}{1}
      {8}{{{\color{numb}8}}}{1}
      {9}{{{\color{numb}9}}}{1}
      {:}{{{\color{punct}{:}}}}{1}
      {,}{{{\color{punct}{,}}}}{1}
      {\{}{{{\color{delim}{\{}}}}{1}
      {\}}{{{\color{delim}{\}}}}}{1}
      {[}{{{\color{delim}{[}}}}{1}
      {]}{{{\color{delim}{]}}}}{1},
}

\title{\textsc{BlendSQL}: A Scalable Dialect for Unifying Hybrid Question Answering in Relational Algebra} 

\author{Parker Glenn, Parag Pravin Dakle, Liang Wang, Preethi Raghavan\\
  Fidelity Investments, AI Center of Excellence\\
  \{\texttt{parker.glenn, paragpravin.dakle,} \\ \texttt{liang.wang, preethi.raghavan\}@fmr.com}}

\begin{document}
\maketitle
\begin{abstract}
Many existing end-to-end systems for hybrid question answering tasks can be boiled down to a ``prompt-and-pray'' paradigm, where the user has limited control and insight into the intermediate reasoning steps used to achieve the final result. Additionally, due to the context size limitation of many transformer-based LLMs, it is often not reasonable to expect that the full structured and unstructured context will fit into a given prompt in a zero-shot setting, let alone a few-shot setting. We introduce \textsc{BlendSQL}, a superset of SQLite to act as a unified dialect for orchestrating reasoning across both unstructured and structured data. For hybrid question answering tasks involving multi-hop reasoning, we encode the full decomposed reasoning roadmap into a single interpretable \textsc{BlendSQL} query. Notably, we show that \textsc{BlendSQL} can scale to massive datasets and improve the performance of end-to-end systems while using 35\% fewer prompt tokens. Our code is available and installable as a package at \href{https://github.com/parkervg/blendsql}{github.com/parkervg/blendsql}.
\end{abstract}

% In doing so, we move away from specialized agent-based architectures and finetuned retrievers, and instead rely on a single uniform few-shot prompt to function over hybrid datasets of diverse size and scope.

\section{Introduction}

% \subsection{Problem Decomposition with LLMs}
\paragraph{Problem Decomposition}
In settings involving both fine-tuning and in-context learning, it has been demonstrated that generating explicit intermediate steps for complex problems can enhance accuracy. In their aptly titled ``Scratchpad'' framework, \citet{nye2021show} revealed that predicting the complete execution trace led to better results when fine-tuning language models to generate execution output. More recent work has shown that prompting LLMs to decompose complicated problems into smaller reasoning steps before generating the final answer can improve results \cite{wei2022chain,yao2023tree,wang2022self}. However, at their core, these methods use natural language as their intermediate reasoning representation. With \textsc{BlendSQL}, we argue that natural language alone is inherently a lossy and ambiguous intermediate reasoning representation. Instead, by decomposing a problem into a SQL-like syntax, we allow for more deterministic reasoning capabilities and better interpretability of intermediate steps by leveraging the compositional nature of relational algebra \cite{codd1970relational}.

\begin{figure}
    \centering
    \includegraphics[width=7.5cm]{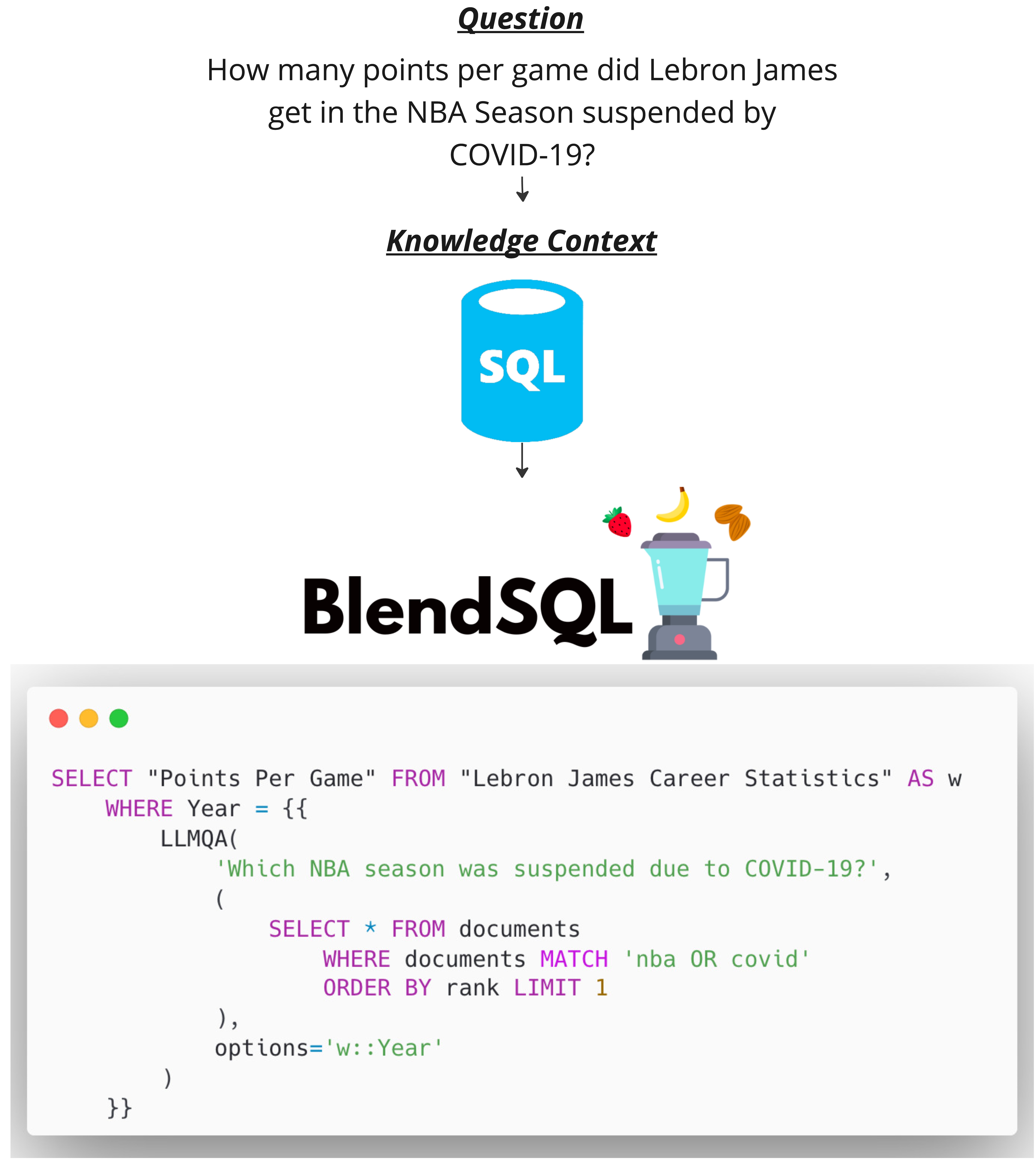}
    \caption{Example \textsc{BlendSQL} representation for an OTT-QA dev example.}
    \label{fig:blendsql_example}
\end{figure}

\paragraph{Agents} Previous works have explored agents: an experts-based framework where the main task is broken down into parts that each agent or expert\footnote{Here an agent or an expert can be anything from a Python script, an API call, a tool or a language model.} carries out \cite{parisi2022talm,codeaspolicies2022,schick2023toolformer,shen2023hugginggpt,liang2023taskmatrix,lu2023gear}. Furthermore, these works are based on the presence of a `routers' or `planners' that understand the given task and break it down into smaller tasks. Although rule-based routers have been widely used, the ability of LLMs to generalize to complex texts has enabled their use as planners. LLMs, as planners, process the input and as output generate text containing the list of subtasks \cite{shen2023hugginggpt}, the agent(s) to use from a given set, what are the inputs to the agent(s) \cite{schick2023toolformer,qin2023tool,Binder}, what is expected as response from the agent(s) \cite{parisi2022talm}. With the scope of the tasks carried out by the agents being very narrow, recent works have primarily focused on the planner component of the framework. For the same, harnessing the in-context learning capabilities of the LLM \cite{shen2023hugginggpt,qin2023tool} or fine-tuning the planner LLM \cite{parisi2022talm,schick2023toolformer,patil2023gorilla,tang2023toolalpaca,wang2023interactive,lu2023gear} have been the two main methods. Additionally, works have used LLMs' in-context learning capability for dataset creation \cite{patil2023gorilla,tang2023toolalpaca}. The parser in \textsc{BlendSQL} acts as a planner and generates the APIs or Ingredients to use in answering the input question. However, it differs from the previous works as it goes beyond generating a single string representation of complicated problem decomposition and simplifies the agent framework using robust and complicated abstract syntax tree (AST) structures to be encoded into a single string. It is also important to note that \textsc{BlendSQL} does not just rewrite the input statement using formal constructs \cite{lu-etal-2021-inter,wu2022autoformalization}, it also adds information using the reference database, input prompt, and internal knowledge of the language models.
%\textcolor{red}{TODO: citations + clarity to this section on agent-based frameworks.} Agent-based frameworks have become increasingly popular as of late. These frameworks typically initialize a grouping of LLM ``agents'', where each agent is specialized at a specific task. While effective, the nature of these agent based frameworks as modular Python scripts leaves room for improvement in terms of reproducibility and clarity of orchestration. 

\paragraph{Text-to-SQL}
On the widely used text-to-SQL dataset Spider  \cite{yu-etal-2018-spider}, many works demonstrate impressive performance in the few-shot or zero-shot setting \cite{dail_sql,pourreza2023din,dong2023c3,liu2023comprehensive}. Even in settings where text-to-SQL is not the focus, existing work details the effectiveness of a common and well-understood intermediate representation like SQL for other reasoning tasks. \citet{hu-etal-2022-context} frames dialogue state tracking (DST) as a text-to-SQL task by encoding domains and slots from MultiWOZ \cite{budzianowski-etal-2018-multiwoz} into a serialized database format and using SQL as an intermediate representation, showing significant improvement over the ``traditional'' key-value style prediction format. 

\paragraph{Hybrid Question Answering}
Unlike text-to-SQL, hybrid question answering involves an implicit decision to access tabular data, unstructured data, or both \cite{chen-etal-2020-hybridqa,zhu-etal-2021-tat,chen2021ottqa}. \citet{li-etal-2021-dual} make this more explicit by routing between end-to-end models and a parser that generates an intermediate SQL representation. Their implementation frames the two as isolated, specialized systems which are unable to pass information to one another. 

For the scope of this work, we define \textit{Hybrid Question Answering} as answering questions using a corpus of tabular and unstructured multi-modal content. While this multi-modal content can encompass text, images, audio, and video, the primary focus of this paper is on unstructured text in conjunction with tabular data. The potential to adapt this process to other formats, such as images, is also discussed in Appendix \ref{app:sec-hybrid-qa-images}. 

% Adding UniK-QA comparison
One prior work, UniK-QA \cite{oguz-etal-2022-unik}, proposes converting structured tabular data into unstructured text data. Our work does the opposite, and instead converts all unstructured text data into tabular data. Empirically, UniK-QA improves on several knowledge-base (KB) QA benchmarks using a KB-to-text preprocessing model. For example, for the KB-to-text model mentioned in the UniK-QA paper, the relationship between different entities is assumed to have a simple triplet representation: <subj, pred, obj> where pred (predicate) defines the relationship from the table name. However, relationships between <subj> and <obj> are often complicatedly embedded in heterogeneous information sources (e.g., audio, documents, or column names of the table), which would then demand more efforts in developing dedicated KB-to-text models. 

% To better utilize the efficiency of structured data whenever possible,
% our method does the opposite by contextualizing the unstructured textual data into necessary structured format.  
\paragraph{Neuro-Symbolic Frameworks}
Most similar to our work is the \textsc{BINDER} method presented in \citet{Binder}, which integrates LLM reasoning into symbolic languages like SQL and Python. While we take inspiration from their neuro-symbolic framework for reasoning, \textsc{BlendSQL} differs in the following ways. (1) We create an API specialized for SQL, enabling reasoning over multi-table databases containing unstructured content, and are capable of executing and optimizing queries containing \texttt{JOIN} statements, aliases, conditional table expressions, and most other SQLite operations (2) We enable constrained decoding according to the database schema (3) We design our API such that users are able to easily create any number of custom functions themselves to use within a \textsc{BlendSQL} script.

\subsection{Main Contributions}
In summary, we make the following contributions. 
\begin{itemize}
\item To the best of our knowledge, we are the first to propose framing the context of hybrid question answering as a relational database.
\item We introduce a new open-source query language, \textsc{BlendSQL}, to orchestrate and optimize hybrid functions across SQL logic and LLM reasoning.
\item We demonstrate that with only a small number of in-context exemplars, \textsc{BlendSQL} can outperform end-to-end methods using 35\% fewer tokens and without direct access to unstructured context.
\end{itemize}

\section{\textsc{BlendSQL}: Overview}
We implement \textsc{BlendSQL} as a superset of the SQLite relational database management system (RDBMS)\footnote{\url{https://www.sqlite.org/}}. An executed \textsc{BlendSQL} script returns a \texttt{smoothie} object as output, containing the final result and intermediate reasoning steps taken. To make a \texttt{smoothie}, we need some ingredients and a blender. We describe these components below.

\subsection{Blender vs. Parser}
We use the term ``blender'' to describe the LLM which receives the prompts used to perform each ingredient function within a \textsc{BlendSQL} script. Prior to execution, the ``parser'' receives a set of few-shot examples and generates a \textsc{BlendSQL} query given a question and database context. We use GPT-4-0613 \cite{achiam2023gpt} as both the blender and the parser for our core experiments.

\subsection{Ingredients}
Ingredients are at the core of a \textsc{BlendSQL} script. They are subprograms used to pass certain logical operations through LLM-based functions, and are denoted by double-curly braces (\texttt{``\{\{''} and \texttt{``\}\}''}).
\textsc{BlendSQL} syntax is represented as a parsing expression grammar, implemented via PyParsing\footnote{\url{https://github.com/pyparsing/pyparsing}} \cite{mcguire2007getting}. Like vanilla SQL, ingredient calls are fully recursive, and the context passed to one can include operations invoking other \textsc{BlendSQL} ingredients.

\begin{figure*}
    \centering
    \includegraphics[width=\textwidth]{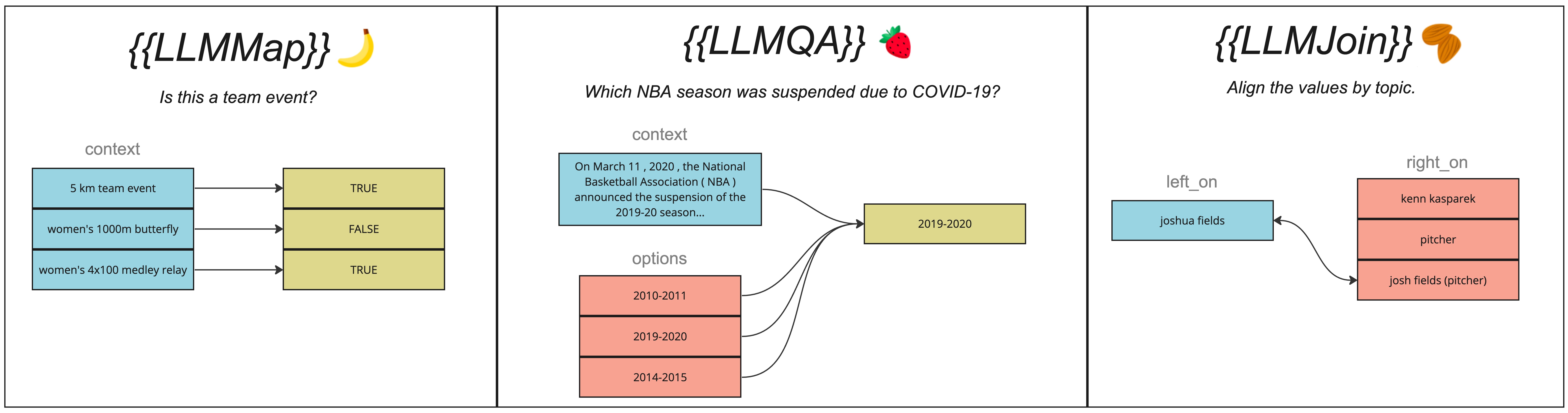}
    \caption{Visualizing built-in \textsc{BlendSQL} ingredients.}
    \label{fig:ingredient_examples}
\end{figure*}

\paragraph{\textit{LLMMap}}
The \textit{LLMMap} ingredient is a unary scalar function, much like \textit{LENGTH} or \textit{ABS} in standard SQLite. The output of this operation is then set as a new column in a temporary table, for later use within the wider query. Specifically, it accepts table and column arguments in the string format ``\textit{table::column}''. It then creates a new column $c'$ with mapped values $v'$, which are the output of the scalar function. With $n_c$ as the number of values in the target column $c$, we get the following.
% https://en.wikipedia.org/wiki/Projection_(relational_algebra)
\begin{align*}
 \mathcal{V} = \{v_i, \forall_{i \in \{1,...,n_c\}}\}
\\
 f_{map}(\mathcal{V}) \rightarrow \{v'_1,...,v'_{n_c}\}
\end{align*}
For example, Figure \ref{fig:ingredient_examples} depicts the \textit{LLMMap} ingredient in action. For each value, an LLM is prompted to answer the question: \textit{``Is this a team event?''} Given the diversity in how team events are denoted in this column (``team event'', ``4x100 medley relay'', etc.), it becomes difficult to write SQLite logic for a generalized solution across the entire table via pure string parsing. \textit{LLMMap}, instead, implements a generalized solution to transform values based on their semantic denotations.

\paragraph{\textit{LLMQA}}
The \textit{LLMQA} ingredient is an aggregate function that transforms a table subset $\mathcal{T}$ into a single scalar value $v'$. 
\begin{align*}
f_{qa}(\mathcal{T}) \rightarrow v'
\end{align*}
As shown in Figure \ref{fig:blendsql_example}, this ingredient may be restricted with the \textit{options} argument, which ensures that the output of the function will be an existing value in a specific column $c$, $v' \in \{v_i, \forall_{i \in \{1,...,n_c\}}\}$\footnote{This constrained decoding is implemented using \href{https://github.com/guidance-ai/guidance}{guidance}, which manipulates the \textit{logit\_bias} argument in the \href{https://platform.openai.com/docs/api-reference/chat/create}{OpenAI API}.}. In Figure \ref{fig:ingredient_examples}, the \textit{LLMQA} function receives a piece of unstructured document context, and the question: \textit{``Which NBA season was suspended due to COVID-19?''} Additionally, it receives a set of column values to use in constraining the language model's generation. 

For fact verification tasks, we implement a modified version of this ingredient, \textit{LLMValidate}, which always returns either a \textit{true} or \textit{false} boolean. 

\paragraph{\textit{LLMJoin}}
The \textit{LLMJoin} ingredient creates a custom mapping between a pair of value sets $V1$ and $V2$. This mapping is then used to create an auxiliary table to carry out a SQL \textit{INNER JOIN} operation.
\begin{align*}
 {f_{join}(\mathcal{V}1, \mathcal{V}2)} \rightarrow \{(v1'_1, v2'_1), ... (v1'_{n_v1}, v2'_{n_v2})\}
\end{align*}

This ingredient is particularly useful in situations when proper foreign keys do not exist, but some semantic alignment is still possible. In Figure \ref{fig:ingredient_examples}, ``joshua fields'' and ``josh fields (pitcher)'' share the same referent. We use the \textit{LLMJoin} ingredient to align the two distinct references, effectively performing a form of entity linking.

\subsection{Query Optimizations}
\begin{figure}
    \centering
    \includegraphics[scale=0.18]{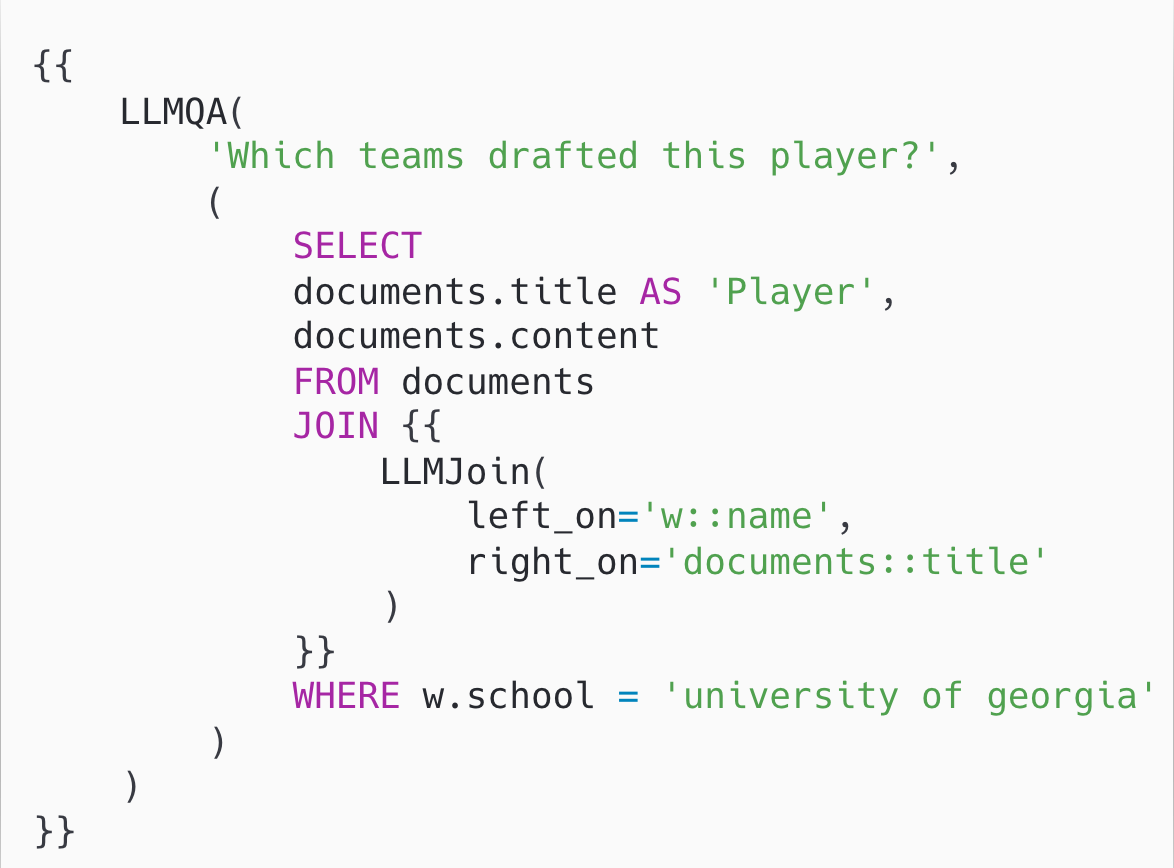}
    \caption{\textsc{BlendSQL} for ``Which teams has the player drafted by the Seattle Mariners in 2008 out of University of Georgia played for in the MLB ?'', aligned with Table \ref{hybridqa_database_example}}
    \label{mariners_example_code}
\end{figure}

When possible, we execute all SQL predicates in a subquery first and assign their outputs to a temporary session table. Since native SQLite operations are relatively inexpensive, we ensure that the expensive LLM-based ingredient functions receive no more (and no less) the subset of data required to generate a faithful execution output. As demonstrated below, we only pass the subset of rows from \textit{w} where \textit{w.school = 'university of georgia'} to the LLM-based ingredients in \ref{mariners_example_code}. We validate these optimizations with an extensive test suite\footnote{\url{https://github.com/parkervg/blendsql/tree/main/tests}}.
\\ 

\begin{table}[]
\resizebox{\columnwidth}{!}{%
\begin{tabular}{@{}lll@{}}
\toprule
\multicolumn{1}{l|}{\textbf{name}}  & \multicolumn{1}{l|}{\textbf{position}}    & \textbf{school}         \\ \midrule
\multicolumn{1}{l|}{joshua fields}  & \multicolumn{1}{l|}{right-handed pitcher} & university of georgia   \\ \midrule
\multicolumn{1}{l|}{dennis raben}   & \multicolumn{1}{l|}{outfielder}           & university of miami     \\ \midrule
\multicolumn{1}{l|}{matt jensen}    & \multicolumn{1}{l|}{second basemen}       & clovis east high school \\ \midrule
\multicolumn{3}{l}{}                                                                                      \\ \midrule
\multicolumn{1}{l|}{\textbf{title}} & \multicolumn{2}{l}{\textbf{content}}                                \\ \midrule
\multicolumn{1}{l|}{kenn kasparek} &
  \multicolumn{2}{l}{\begin{tabular}[c]{@{}l@{}}kenn anthony kasparek \\ ( born 1985-9-23 ) is...\end{tabular}} \\ \midrule
\multicolumn{1}{l|}{pitcher} &
  \multicolumn{2}{l}{\begin{tabular}[c]{@{}l@{}}in baseball , the pitcher is the \\ player who throws the baseball from...\end{tabular}} \\ \midrule
\multicolumn{1}{l|}{josh fields (pitcher)} &
  \multicolumn{2}{l}{\begin{tabular}[c]{@{}l@{}}joshua david fields ( born 1985-8-19 ) \\ is an american professional baseball pitcher...\end{tabular}} \\ \bottomrule
\end{tabular}%
}
\caption{Example database configuration from a HybridQA example, aligned with \ref{mariners_example_code}. The top table \textit{w} contains the structured data corresponding to a table found on a Wikipedia page, while the bottom table \textit{documents} contains unstructured data found in the article content of a Wikipedia page, indexed with FTS5\footnote{\url{https://www.sqlite.org/fts5.html}}.}
\label{hybridqa_database_example}
\end{table}

% \begin{lstlisting}[language=SQL,label={lst:example-blendsql},title={\textsc{BlendSQL} for ``Which teams has the player drafted by the Seattle Mariners in 2008 out of University of Georgia played for in the MLB ?'', aligned with Table \ref{hybridqa_database_example}},breaklines=true,showstringspaces=false]
% {{
%     LLMQA(
%         'Which teams drafted this player?',
%         (
%             SELECT documents.title AS 'Player', documents.content FROM documents JOIN {{
%                 LLMJoin(
%                     left_on='w::name',
%                     right_on='documents::title'
%                 )
%             }}
%             WHERE w.school = 'university of georgia'
%         )
%     )
% }}
% \end{lstlisting}

\section{Datasets}
Due to the joint usage of SQL-like syntax and LLM-based ingredients, we expect our \textsc{BlendSQL} to perform well on hybrid question answering datasets. To validate our hypothesis, we experimented on several challenging benchmark datasets requiring reasoning over both structured and unstructured knowledge.

\subsection{HybridQA}
Compared to previous QA tasks utilizing only passages (unstructured) or tabular information (structured), HybridQA contains challenging questions whose answers demand heterogeneous forms of information collected from Wikipedia tables and passages \cite{chen-etal-2020-hybridqa}. In an ablation study, the original authors show that a table-only model achieves an accuracy of 8.4 on the dev set, and a passage-only model achieves an accuracy of 19.5. Combining both sources, their baseline model improves drastically to 44.0, highlighting the importance of a model adept at combining both hybrid data sources. 

\subsection{OTT-QA}
The Open Table-and-Text Question Answering dataset (OTT-QA) is a ``decontextualized'' variant of HybridQA, which requires both text and table retrieval over a large corpus \cite{chen2021ottqa}. On the OTT-QA dataset, we demonstrate the potential of \textsc{BlendSQL} to act as both the retriever and reader within a unified query language. In the multi-hop reasoning OTT-QA requires, the relevant passages can only be found after completing some operation on the structured tables, and vice versa. We enable this communication between the reader and retriever components via BM25 full-text search enabled with the built-in FTS5 extension\footnote{\url{https://www.sqlite.org/fts5.html}} in SQLite. For example, in Figure \ref{fig:blendsql_example}, we first retrieve unstructured text from the \textit{documents} table with the highest BM25 relevancy to the query \textit{``nba OR covid''}. Then, given the structure of the table \textit{``Lebron James Career Statistics''}, we constrain the possible generations given our unstructured context to only a value occurring in the \textit{``Year''} column.

\subsection{FEVEROUS}
The FEVEROUS (Fact Extraction and VERification Over Unstructured and Structured information) dataset contains claims accompanied by context sentences and tables from Wikipedia \cite{aly-etal-2021-fact}. Each claim is classified as either ``supports'', ``refutes'', or ``not enough info''. By using \textsc{BlendSQL} as an intermediate representation, we are able to frame the notion of ``truth'' as a function over facts found within a database. Additionally, by generating an intermediate representation of the FEVEROUS claim, we not only produce an interpretable decomposition of the implicit truth claims, but we also have a reusable blueprint for future fact verification even if the values in our underlying database are updated. This differs from the traditional end-to-end approach, where the full hybrid context must be passed each time a prediction is made.

\paragraph{\textsc{BlendSQL} as Predicate Logic}
\label{sec:blendsql_predicate_logic}
\textsc{BlendSQL} transforms the propositional claims of FEVEROUS into predicate logic, providing a new language to evaluate the truth value of a statement given world knowledge in a relational database. For example, we can take the following (abbreviated) example from the FEVEROUS dev set.

\begin{quote}
\textit{Pesamino Victor (an association footballer) and his team lost in all their international matches.}
\end{quote}
Given the constant $p$ for Pesamino Victor and two-place predicates $\textsc{PlaysOn}$, $\textsc{PlayedBy}$, and $\textsc{Won}$, this becomes the following.
\begin{flalign*}
\exists{t}(\textsc{AssociationTeam}'(t) \wedge \textsc{PlaysOn}'(p, t))
\\\wedge \neg \exists{m}(\textsc{InternationalMatch}'(m)\\\wedge \textsc{PlayedBy}'(t, m) \wedge \textsc{Won}'(t, m))
\end{flalign*}

Below, we display the \textsc{BlendSQL} program corresponding to the same statement, given the underlying structured (the \textit{Pesamino Victor} table) and unstructured (\textit{documents}) context.

% \begin{lstlisting}[language=SQL,caption={},breaklines=true,showstringspaces=false]
% SELECT (
%     {{
%         LLMValidate(
%             'Is Pesamino Victor an 
%             association footballer?',
%             (
%                 SELECT * FROM documents 
%                 WHERE title = 'pesamino victor'
%         ) 
%     }}
%     ) AND NOT EXISTS (
%         SELECT * FROM "Pesamino Victor" AS w 
%         WHERE {{LLMMap('Did they win this match?', 'w::opposition (result)')}} = TRUE
%     )
% \end{lstlisting}

\begin{figure}[h]
    \centering
    \includegraphics[scale=0.18]{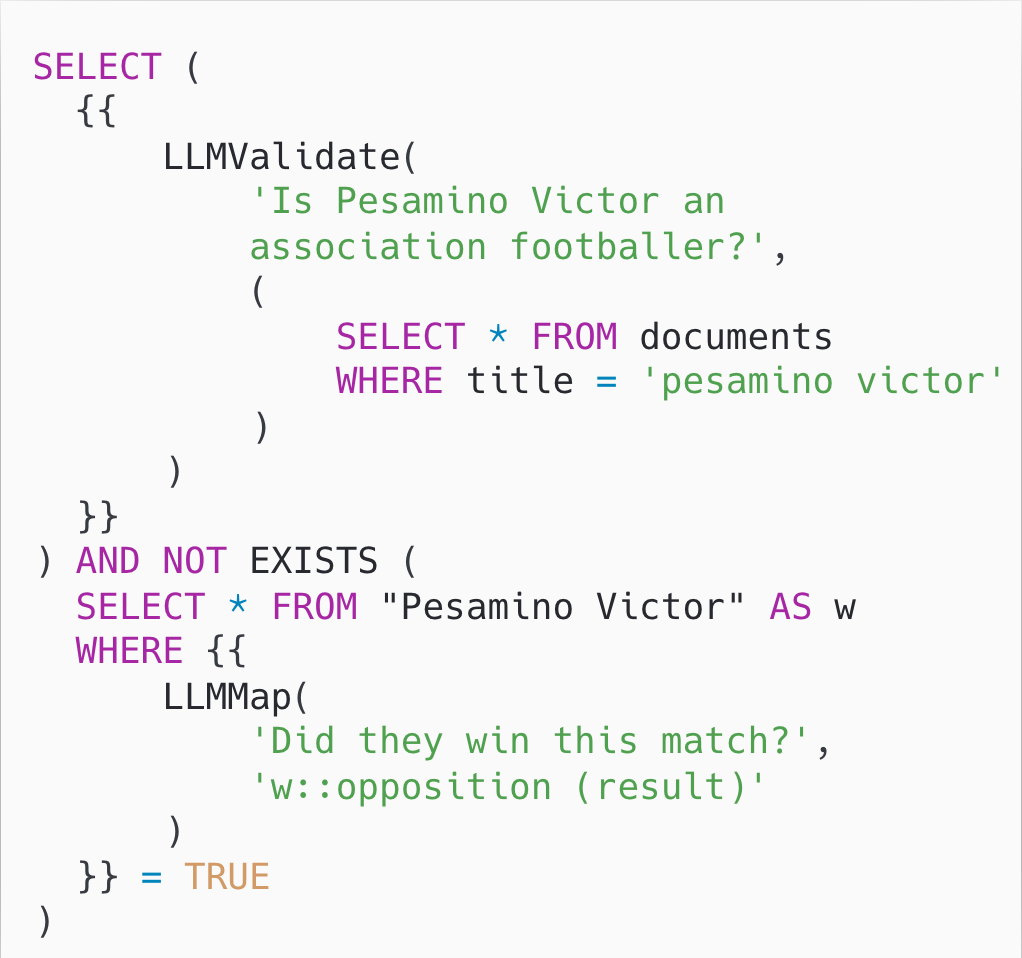}
\end{figure}

\section{Experiments}

\subsection{Dataset Pre-processing}
In all the datasets we evaluate, we place both structured and unstructured contexts into a SQLite database as shown in Table \ref{hybridqa_database_example}. We use the FTS5 extension\footnote{\url{https://www.sqlite.org/fts5.html}} to create a full-text index with BM25 ranking over the \textit{title} and \textit{content} columns in our \textit{documents} table.

Additionally, we use the fuzzy string matching described in \citet{LinRX2020:BRIDGE} to align a given question to relevant values in the underlying database, and provide these ``bridge'' matches as hints in our prompt. We serialize our databases as code, as proven effective for text-to-SQL in the study by \citet{nan-etal-2023-enhancing}. Specifically, we include the \textit{CREATE} clause for each table in the database, along with $n$ example rows for every table except \textit{documents}.
% \footnote{For a full description of hyperparameters for each dataset, see Appendix \ref{sec:dataset-hyperparameters}}
In many datasets, the unstructured text includes lengthy passages that are often irrelevant to the given question. By withholding the unstructured content from the parser, we are able to scale to massive datasets such as OTT-QA.

\paragraph{End-to-End Prompting on HybridQA}
On the HybridQA dataset, many data points contain context data that is too large to fit into a single GPT-4 prompt. To solve for this, we include all tabular data, and truncate the values of the unstructured content to a max of 400 characters.

\subsection{Teaching \textsc{BlendSQL} via In-Context Learning}
We implement our few-shot prompting via the guidance toolkit\footnote{\url{https://github.com/guidance-ai/guidance/tree/0.0.64}}, which supports handlebars-style syntax to control the prompting workflow\footnote{For the full prompt, see Appendix \ref{sec:few-shot-prompt}}. For each dataset, we randomly sample $\sim$10 examples from the train set and annotate their \textsc{BlendSQL} representation. For HybridQA and FEVEROUS, we use 12 examples; for OTT-QA, we use 9.

\subsection{Metrics}
In an attempt to measure the true performance of our approach without overfitting to spurious answer formats present in the free-text question-answering datasets, we employ the denotation accuracy metric used by \citet{Binder}, from the script \href{https://github.com/xlang-ai/Binder/blob/95f2facc45910959812d625e501edfa9f889abd9/utils/normalizer.py}{here}. This metrics measures semantic ``denotation accuracy'', normalizing different output formats (such as ``seven'' and ``7''). On the HybridQA dev set, for example, 101 gold answers contain natural language numbers (\textit{five million}, \textit{three campuses}), and 697 gold answers contain numeric representations of numbers (\textit{4 million}, \textit{6}). The official HybridQA evaluator would judge these different answer formats as incorrect, and therefore, we opt for measuring semantic denotation instead.

\section{Results}
We direct the curious reader to Table \ref{example_predictions} in the Appendix for example predictions and more analysis.

\begin{table*}
\begin{minipage}[t]{\columnwidth}
\centering
\begin{tabular}{@{}ll@{}}
\toprule
\textbf{Method}                                                      & \textbf{Accuracy} \\ \midrule
\multicolumn{1}{c}{\textit{Oracle Document Retriever}}                          &                   \\
End-to-End \cite{sui2023gpt4table}                                                            & 56.68             \\ \midrule
\multicolumn{1}{c}{\textit{Searching Entire Context}} &                   \\
End-to-End                                                           & 49.13             \\
\textsc{BlendSQL}                                                             & 52.89             \\
\textsc{BlendSQL} + End-to-End                                                & \textbf{57.76}    \\ \bottomrule
\end{tabular}
\caption{HybridQA dev set results using GPT-4-0613. \textit{End-to-End} methods are zero-shot, whereas \textsc{BlendSQL} uses 12 few-shot exemplars to teach the model this new SQL dialect as an intermediate representation.}
\label{hybridqa_results}
\end{minipage}\hfill 
\begin{minipage}[t]{\columnwidth}
\centering
\begin{tabular}{@{}ll@{}}
\toprule
\textbf{Method}                         & \textbf{Accuracy} \\ \midrule
\textit{Transformer-based Doc Retriever} &                   \\
FR+CBR \cite{chen2021ottqa}                               & 28.1            \\
CARP \cite{zhong2022reasoning}                                   & 33.2              \\
CORE \cite{ma-etal-2022-open-domain}                                    & 49.0              \\ \midrule
\textit{BM25 Doc Retriever}        &                   \\
\textsc{BlendSQL$^*$}                                & 34.15             \\ \bottomrule
\end{tabular}%

\caption{OTT-QA dev set results using GPT-4-0613. The previous systems are all finetuned, using dedicated transformer-based retriever components for both the 400k tables and 5M passages. \textsc{BlendSQL} uses the top-4 table retriever predictions from \citet{ma-etal-2022-open-domain}, but encodes both the BM25 passage retrieval + reader steps in a unified program.}
\label{ottqa-results}
\end{minipage}
\end{table*}

\begin{table}[]
\begin{tabular}{ll}
\hline
\textbf{Method}         & \textbf{Accuracy} \\ \hline
End-to-End \cite{sui2023gpt4table}              & \textbf{83.21}    \\
\textsc{BlendSQL (3 rows)}$^\ddag$       & 65.7              \\
\textsc{BlendSQL} (Entire table)$^\ddag$ & 68.03             \\ \hline
\end{tabular}
\caption{FEVEROUS dev set results using GPT-4-0613. }
\label{feverous-results}
\end{table}

% \section{Discussion}

\subsection{HybridQA}
Unlike in intermediate representations based on natural language, a \textsc{BlendSQL} script may result in an execution error and fail to produce a response. Additionally, the returned \textit{LLMQA} ingredient may deem the question as unanswerable given the provided context and respond with something in the spirit of ``This table does not provide the necessary info.'' In these cases, we experiment with falling back to the end-to-end prompting style as a last resort, as denoted by ``BlendSQL + End-to-End'' in Table \ref{hybridqa_results}.

With 12 few shot exemplars, the parsed \textsc{BlendSQL} query was unable to generate a prediction on 33\% (1,173 examples out of 3,466) of the dev set. Even with this shortcoming, it outperformed the truncated context, end-to-end method by 3.76\%. By using \textsc{BlendSQL} first, and falling back to end-to-end prompting, we boost performance by 8.63\% to \textbf{57.76\%}, outperforming the oracle document retriever approach of \citet{sui2023gpt4table}. Notably, on the subset of 2,293 questions that generated a valid \textsc{BlendSQL} script, we reached an accuracy of 64.43\%, suggesting that improved few-shot examples and perhaps more refined ingredients may have the potential to further boost performance.

\subsection{OTT-QA}
Depicted in Table \ref{ottqa-results}, \textsc{BlendSQL} allows for competitive performance among existing finetuned benchmarks, with only 9 few-shot examples. Even more promising is \textsc{BlendSQL}'s use of a naive BM25 document retriever, and ability to encode the unified reasoning roadmap into a single, debug-gable query. While we use the predictions from the table retriever described in \citet{ma-etal-2022-open-domain}, it is up to our parser to write a satisfactory FTS5 query for retrieving documents from the BM25 index store. For these reasons (in-context learning, with a naive BM25 retriever), it is difficult to compare our results to existing literature directly. Future work involving a more robust retrieval algorithm (e.g., retrieval with vector embeddings) has the potential to further improve results.
\footnotetext{%
$^\ddag$ We follow the previous work in \citet{sui2023gpt4table} and only evaluate on claims which include table-based evidence. However, we discard data points with the label ``not enough enough'', making the results not directly comparable.

$^*$ To normalize different output formats (such as ``seven'' and ``7''), we employ the denotation accuracy metric used by \citet{pasupat-liang-2015-compositional}, from the script \href{https://github.com/ppasupat/WikiTableQuestions/blob/9445de5efd1041a2576a065f1b0320385975f917/evaluator.py}{here}. It is unclear if the previous methods use the same evaluation script.
}
\subsection{FEVEROUS}
As shown in Table \ref{feverous-results}, \textsc{BlendSQL} does not outperform the end-to-end prompting style on the FEVEROUS dataset. As described in \ref{sec:blendsql_predicate_logic}, this is likely due to the complexity and ambiguity of outlining predicate logic to evaluate the truth value of a given claim, compared to the other hybrid QA tasks. Not only have we prompted our parser to compose a script that evaluates to the correct judgment (``supports'' or ``refutes''), but we have implicitly asked it to identify the various atomic truth claims made within a claim, and set boundaries for each with respect to database context. Additionally, the table structures in FEVEROUS tend to deviate far from the traditional relational model, with many subtables and empty values.

% of which the underlying data might be in irregular format or simply be missing. 

% In a way, for hybrid qa, ``the end justifies the means'', which is less true for translating claims to a predicate logic-style notation.

\section{Error Analysis}
\label{sec:error_analysis}
To better understand the advantages and limitations of \textsc{BlendSQL}, we annotate randomly sampled 50 \textsc{BlendSQL} question-answer pairs from HybridQA mistakes.
Our chosen denotation accuracy metric judges each datapoint as either a 1 or 0. Out of the 3,466 datapoints in the HybridQA dev set, we see 1,464 with an accuracy score of 0. 
% The prediction mistake is defined as predictions with zero EM score, which totaled 1,464 as it is compared to the original 3,466 HybridQA dev set. 
Although the number 1,464 seems intimidating, it is important to remember that we achieved a literature-comparable score of 57.76 and to realize that there are many false negatives from those errors. To distinguish those false negative errors, we describe \textbf{Annotation Categories} as the parent categories containing false negative errors and the \textbf{BlendSQL Error Categories} as the child categories containing only true negative errors.

\begin{figure}
    \centering
    \includegraphics[scale=0.45]{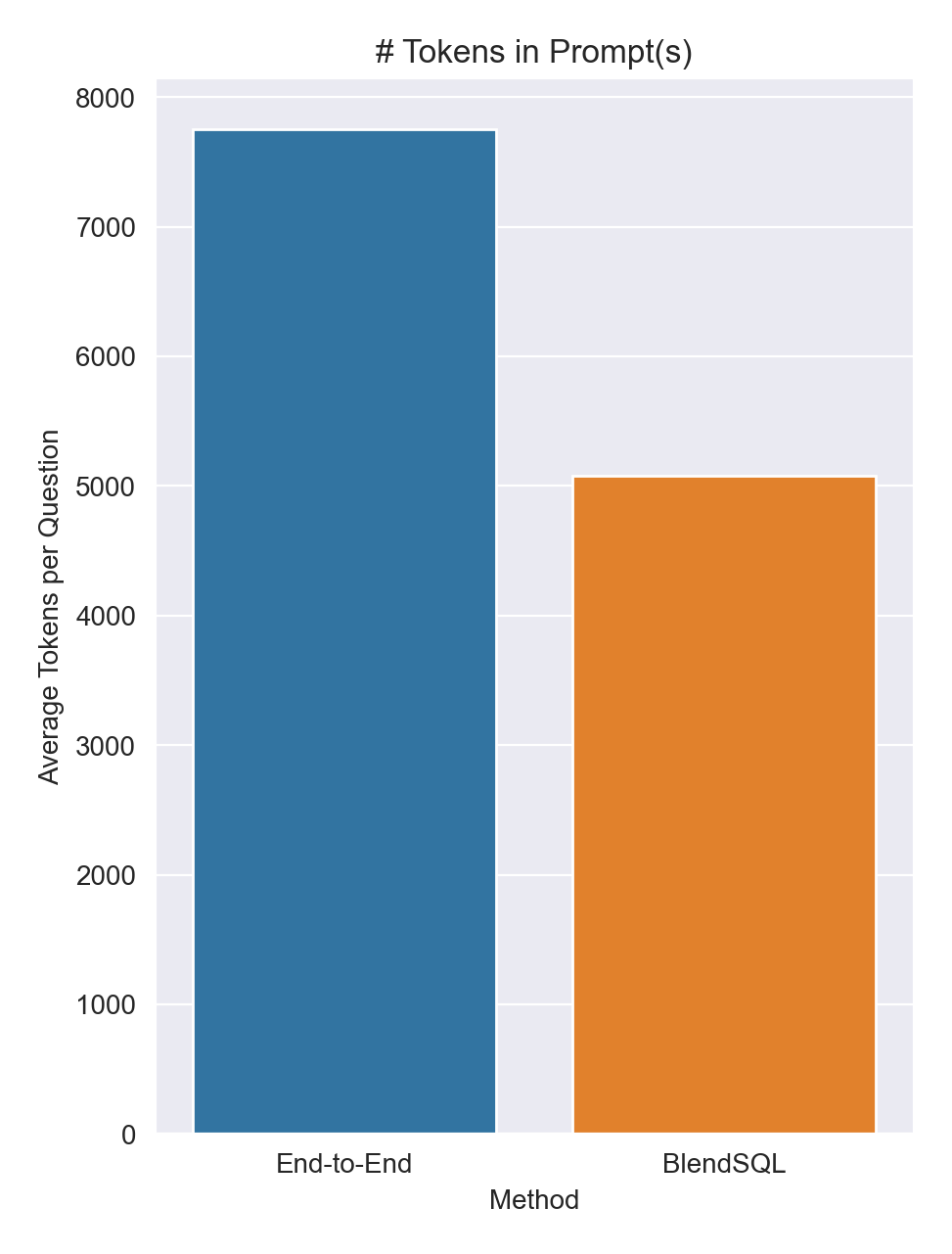}
    \caption{Average prompt tokens per question on the HybridQA dev set. \textsc{BlendSQL} enables efficient filtering of large context databases to decrease data passed to the LLM by 35\%.}
    \label{fig:prompt_token_counts}
\end{figure}

\begin{figure*}
    \centering
    \includegraphics[width=\textwidth]{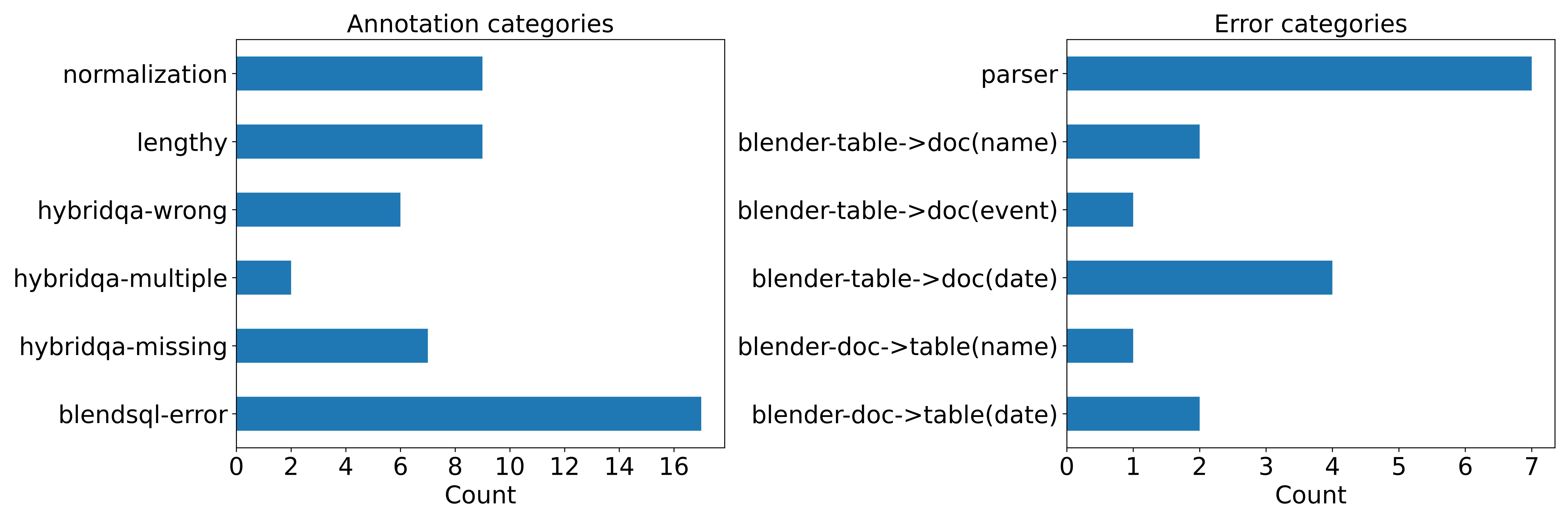}
    \caption{Error analysis on a random 50 samples of the HybridQA dev set. As described in Section~\ref{subsec:annotat_cateogies},
    left shows 17 (34\%) of the error are True Negative Errors for BlendSQL (\textit{blendsql-error}).
    Right shows the causes of those True Negative Errors.}
    \label{fig:error_categories}
\end{figure*}

\subsection{Annotation Categories} 
\label{subsec:annotat_cateogies}
Figure~\ref{fig:error_categories} shows an overview of the annotation categories, among the 50 mistakes, there are considerable amount of mistakes are false negatives because those \textsc{BlendSQL} answers are semantically equivalent to the HybridQA's ground-truth answer. The annotation categories, along with their definitions, are described below:

\paragraph{Lengthy}
Our \textit{LLMQA} ingredient tends to provide a more elaborate answer instead of naively an address, name, or date. These elaborate but semantically equivalent answers are annotated as \textsc{lengthy}. 

\paragraph{Normalization}
Even with the normalization script of \citet{Binder}, some unwanted formatting variations remain. This second source of false negative is due to text normalization such as "\$20 millions" versus "20,000,000", "Belize City" versus "belize" and "METREX" versus "Metrex Network". We annotate this second source of false negative as \textsc{normalization}. 

\paragraph{HybridQA-Centric}
There are also some ground-truth answers from HybridQA that we believe are partially wrong, ambiguous with multiple answers, or require additional background information not present in the document or the table. We annotate those observations into \textsc{HybridQA-wrong}, \textsc{HybridQA-multiple}, and \textsc{HybridQA-missing}. 

\paragraph{BlendSQL}
The mistake caused by \textsc{BlendSQL} only occupies 17 out of the 50 sampled mistakes, amounting to 34\% of the true negative rate. 

\subsection{BlendSQL Error Categories}
\label{subsec:blendsql_error_categories}
Focusing on those 17 true negative \textsc{BlendSQL} mistakes, we further break down the error into 6 major categories with Figure~\ref{fig:error_categories}.

\paragraph{Parser} The \textsc{parser} category indicates the incorrect answer is caused by a mistake in \textsc{BlendSQL} syntax. For example, the ``greater than'' syntax (\textit{WHERE "date" > \{\{LLMQA(...)\}\}}) is incorrectly generated as ``equals'' (\textit{WHERE "date" = \{\{LLMQA(...)\}\}}). 

\paragraph{Blender-Centric} Besides the most common \textsc{parser} error, we see five additional categories indicating the direction of the multi-hop reasoning and the topic of the mistakes. Take the following question as an example. 
\begin{quote}
\textit{What is the difference in time between Jose Reliegos of Spain and the person born 5 September 1892 who competed at the 1928 Olympics?} 
\end{quote}

The answer requires a full-text search of the athlete name within the document via \textit{documents MATCH 'born 5 september 1892'}, then a SQL query to calculate the time difference based upon the searched athlete name. In this given example, the error is annotated as \textsc{Blender-doc->table(date)}. 

\subsection{Error Causes} A majority of the \textsc{blensql} mistakes came from the parser generation. We find that sometimes the generated \textsc{Blendsql} syntax uses \textit{LLMQA} to answer questions that are better suited for the \textit{LLMJoin} operation. Those inappropriate parses hence incur incorrect answers related to date and numerical operation. 
% The good news is that this limitation can be improved by providing more instructive a-few-shot \textsc{BlendSQL} syntax examples, that is, we can inject more date and numerical related questions to \textit{LLMJoin} instead of \textit{LLMQA}. 

Among those blender mistakes, we find the multi-hop reasoning from \textsc{doc->table} is usually more challenging than \textsc{table->doc}. The reason is that searching for a potential filter over an unstructured document is more difficult and open-ended than a column filter over a structured table.

\section{Open Model Evaluations}
Closed models like GPT-4-0613, while powerful, lack a level of transparency and interpretability that is valuable to the research community. To this end, we also evaluate two open-source language models built for coding tasks: DeepSeek-Coder-6.7b-Instruct\footnote{\url{https://huggingface.co/deepseek-ai/deepseek-coder-6.7b-instruct}} \cite{guo2024deepseek} and StarCoder2-15b\footnote{\url{https://huggingface.co/bigcode/starcoder2-15b}} \cite{lozhkov2024starcoder}. Both models are autoregressive decoders with a context window of 16k tokens. We run experiments on the HybridQA dataset, and use each respective model as both the parser and blender to best reproduce the core experiments with GPT-4-0613.

\subsection{Open Model Error Analysis}
\begin{figure}
    \centering
    \includegraphics[scale=0.25]{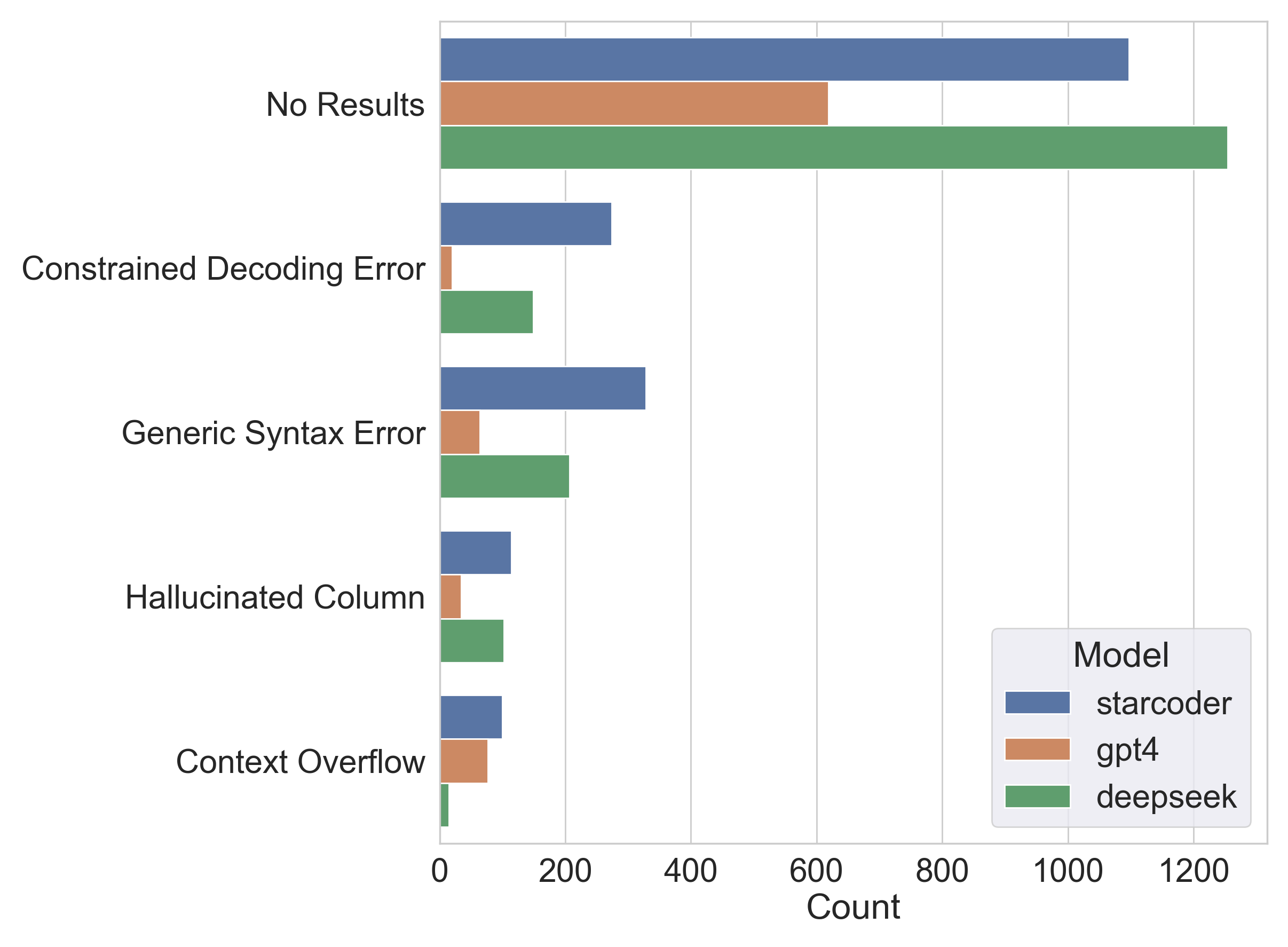}
    \caption{Detailing specific execution errors on the HyrbidQA dev set with various models as the parser. Most commonly, the \textit{LLMQA} subprogram receives an empty table (e.g., due to an overly specific FTS5 query) and fails to produce a response.}
    \label{fig:open_model_syntax_errors}
\end{figure}

As shown in Table \ref{open-model-results}, the two open models fail to match the performance of GPT-4-0613. This can be partly attributed to the ability of the parser to generate valid \textsc{BlendSQL} syntax: while GPT-4-0613 only generated bad syntax on 3\% of the HybridQA dev set, DeepSeek-Coder and StarCoder2 saw rates of 13\% and 21\%, respectively. We explore specific execution errors yielded by different parser models in Figure \ref{fig:open_model_syntax_errors}. In this Figure, we consider everything to the right of ``No Results'' to be a syntax error. Referring back to section \ref{subsec:blendsql_error_categories}, these errors are subsets of the \textsc{Parser} category.

\begin{table}[]
\centering
\resizebox{\columnwidth}{!}{%
\begin{tabular}{@{}ll|l|l@{}}
\toprule
\textbf{Model}                                    & \textbf{Accuracy} & \textbf{F1} & \textbf{\% Bad Syntax} \\ \midrule
\multicolumn{1}{l|}{DeepSeek-Coder-6.7b-Instruct} & 26.90             & 14.05       & 0.13                         \\ \midrule
\multicolumn{1}{l|}{StarCoder2-15b}               & 26.23             & 11.53       & 0.21                         \\ \midrule
\multicolumn{1}{l|}{GPT-4-0613}                        & 52.89             & 45.02       & 0.03                         \\ \bottomrule
\end{tabular}%
}
\caption{Comparing open code-finetuned LLMs against GPT-4-0613 on the HybridQA dev set. Despite its relatively small size, DeepSeek-Coder-6.7b-Instruct outperforms StarCoder2-15b. Both open models, however, perform far below GPT-4-0613.}
\label{open-model-results}
\end{table}

\section{Future Work}
We hope to study the ability of LLMs not only to use functions from a previously unseen SQL dialect but also to create functions that generalize certain logical patterns, as described in \citet{cai2023large}. Additionally, while instruction fine-tuned LLMs show impressive abilities in executing \textsc{BlendSQL} scripts, we hope to experiment with more specialized and inexpensive models for modular tasks such as fact verification and DPR retrieval.

Finally, as many errors came from \textsc{BlendSQL} scripts that were unable to execute properly, we hope to explore hybrid QA tasks as a form of interactive semantic parsing \cite{Elgohary20Speak,glenn-etal-2023-correcting}.

% Finally, the compositional nature of SQL leaves room for more aggressive caching of subqueries in \textsc{BlendSQL}. Currently, we implement a local cache within the scope of the current query. A database-oriented cache, however, would be able to benefit from frequently queried subproblems within a common context. For example, if the subquery \textit{SELECT * FROM w WHERE \{\{LLMMap('What was the total score?', 'column::score')\}\} > 3} was frequently called, we would cache its result to avoid resubmitting repetitive queries to the same database context, further decreasing the tokens passed as shown in Figure \ref{fig:prompt_token_counts}.

\section{Conclusion}
We introduce \textsc{BlendSQL}, a scalable dialect for problem decomposition and hybrid question-answering. Results show competitive performance on popular benchmarks while using only $\sim$10 few-shot examples. Additionally, on the HybridQA dataset, we improve the performance of a naive end-to-end system by 8.63\%, while using 45\% fewer prompt tokens. We open-source all code and present an installable Python package for future researchers to further explore \textsc{BlendSQL} as an intermediate representation for hybrid question-answering.

\section{Limitations}
As our core experiments were conducted using GPT-4-0613, the API costs associated with reproducing the experiments may be a limiting factor. In an attempt to remedy this, we make our work open source and share all evaluation outputs. 

Additionally, while we show that our approach can minimize prompt tokens and improve performance, we require an extra step of pre-processing all context into a SQLite database. We aim to streamline this pre-processing by adding support for more database management systems, and creating helper scripts for transforming a hybrid question-answering context into the appropriate database format.  
\bibliography{acl2023}
\bibliographystyle{acl_natbib}
\appendix
\section{Appendix}
\subsection{Inferring Outputs from SQL Syntax}
Ingredients in \textsc{BlendSQL} are inherently stateless; they do not receive any external information other than that which is explicitly passed in as arguments. This approach leads to a view of problem decomposition where each ingredient call is only responsible for the specialized task it's been setup to solve for. 

However, oftentimes, an ingredient's placement within the larger SQL syntax can provide useful signals for the downstream LLMs. For example, consider the \textsc{BlendSQL} query below.

\begin{figure*}[b!]
    \centering
    \includegraphics[width=\linewidth]{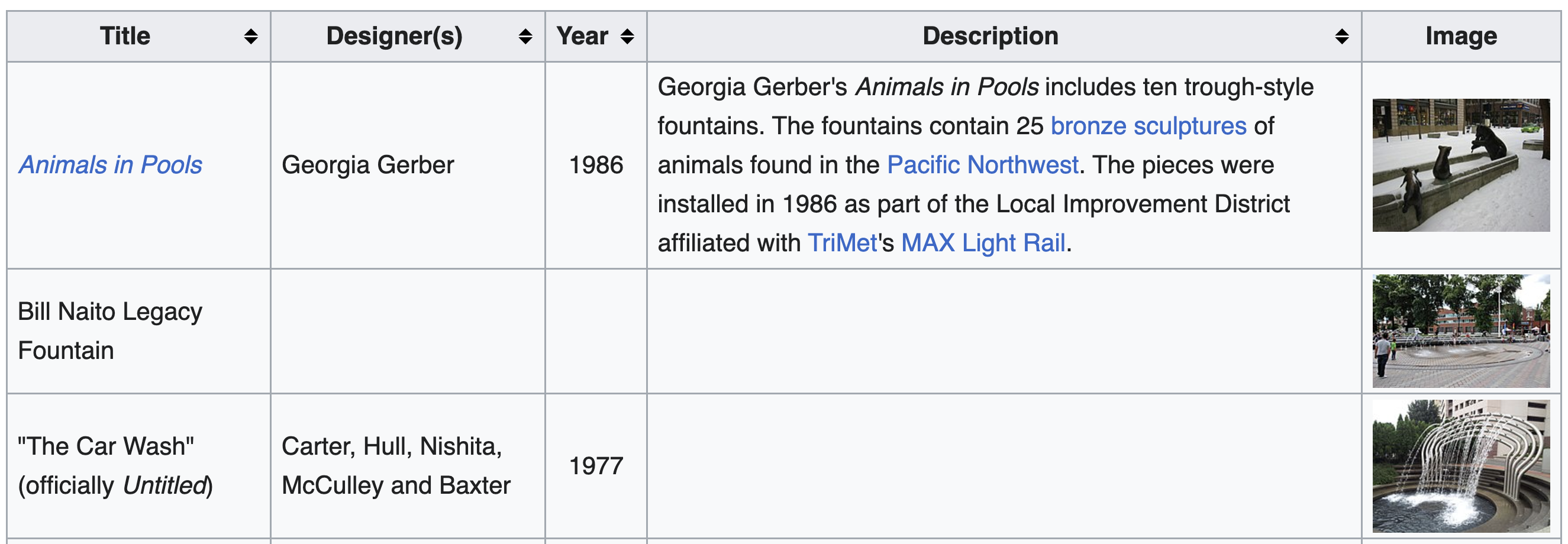}
    \caption{Sample of the hybrid table from the \textit{Fountains in Portland, Oregon} Wikipedia page used for demonstrating Visual Question Answering using BlendSQL}
    \label{app_fig:vqa_table}
\end{figure*}

\begin{lstlisting}[language=SQL,breaklines=true,showstringspaces=false]
SELECT * FROM w WHERE {{
    LLMMap(
        'What state is this city in?', 
        'w::city'
}} = 'CA'
\end{lstlisting}

We have a single \texttt{LLMMap} ingredient call, which will receive all distinct values from the \texttt{city} column. Examining the query, we also understand that we expect a returned string that looks like 'CA', as opposed to 'California', or any other equivalent denotation. By traversing the AST and extracting the arguments of predicates like \texttt{=}, \texttt{>}, and \texttt{<}, we are able to inject this prior knowledge into the \texttt{LLMMap} call with a simple flag \textit{``Here is an example output: CA''}.
Additionally, by utilizing open models with publicly accesible logits, we are able to directly enforce string patterns at the decoding level \cite{willard2023efficient}. For example, take the below query.

\begin{lstlisting}[language=SQL,breaklines=true,showstringspaces=false]
SELECT * FROM w WHERE {{
    LLMMap(
        'Is this a team event?', 
        'w::event'
}} = TRUE
\end{lstlisting}

By using the same AST traversal logic described above, we can infer that we expect the \texttt{LLMMap} to return a boolean datatype. Using constrained generation, we are able to only produce a response from a language model that matches the regular expression \texttt{((true|false);)+}\footnote{We use a semicolon as a separator here}.

\section{Few-Shot Prompt}
\label{sec:few-shot-prompt}
Below we show the few-shot prompt we used for our core experiments, along with an example database serialized in the ``code'' format.

\begin{lstlisting}[language=json,caption={},breaklines=true,showstringspaces=false]
{{#system~}}
Generate BlendSQL given the question to answer the question correctly.
BlendSQL is a superset of SQLite, which adds external function calls for information not found within native SQLite.
These external functions should be wrapped in double curly brackets.

If question-relevant column(s) contents are not suitable for SQL comparisons or calculations, map it to a new column using the new function:
    `LLMMap('question', '{table}::{column}')`

If mapping to a new column still cannot answer the question with valid SQL, turn to an end-to-end solution using the new function:
    `LLMQA('{question}', ({blendsql}))`
    
If we need to do a `join` operation where there is imperfect alignment between table values, use the new function:
    `LLMJoin(({blendsql}), options='{table}::{column}')`

ONLY use these BlendSQL functions if necessary. 
Answer parts of the question in vanilla SQL, if possible.
{{~/system}}

{{#user~}}
{{few_shot_examples}}

{{serialized_db}}
Question: {{question}}
BlendSQL:
{{~/user}}

{{#assistant~}}
{{gen "result" temperature=0.0}}
{{~/assistant}}
\end{lstlisting}

\label{sec:example-serialized-db}
\begin{lstlisting}[language=json,breaklines=true,showstringspaces=false]
CREATE TABLE "w" (
"index" INTEGER,
  "no" INTEGER,
  "rider" TEXT,
  "team" TEXT,
  "motorcycle" TEXT
)
/*
3 example rows:
SELECT * FROM w LIMIT 3
 index  no          rider                 team      motorcycle
     0   1   carl fogarty   ducati performance      ducati 996
     1   4 akira yanagawa kawasaki racing team kawasaki zx-7rr
     2   5  colin edwards        castrol honda      honda rc45
*/

CREATE VIRTUAL TABLE "documents" USING fts5(title, content, tokenize = 'trigram')
\end{lstlisting}

% \section{Dataset Hyperparameters}
% \label{sec:dataset-hyperparameters}

% \begin{table*}[]
% \centering
% \begin{tabular}{|l|p{3em}|p{3em}|p{5em}|p{5em}|l|p{5em}|}
% \hline
% \textbf{Dataset} &
%   \textbf{Base Model} &
%   \textbf{Temp.} &
%   \textbf{\# Few Shot Examples} &
%   \textbf{Serialized DB Format} &
%   \textbf{\begin{tabular}[c]{@{}l@{}}Max \# Rows \\ Shown per Table\end{tabular}} &
%   \textbf{Content Truncation Limit} \\ \hline
% HybridQA & GPT-4 & 0.0 & 12 & code & 3 & 400  \\ \hline
% OTT-QA   & GPT-4 & 0.0 & 9  & code & 5 & N.A. \\ \hline
% FEVEROUS & GPT-4 & 0.0 & 12 & code & 3 & N.A. \\ \hline
% \end{tabular}
% \caption{Hyperparameters used for various dataset evaluations. ``code'' serialized DB format refers to the format shown in \ref{sec:example-serialized-db}. ``Content Truncation Limit'' is the maximum number of characters we allow to be serialized in the \textit{content} column of the \textit{documents} table when doing end-to-end inference.}
% \end{table*}

\section{Hybrid QA over Images}
\label{app:sec-hybrid-qa-images}

The main paper discusses the Hybrid Question Answering task with text as the unstructured media in addition to tabular data. The methodology used for text can easily be extended for images for Visual Question Answering (VQA) task\footnote{\href{https://github.com/parkervg/blendsql/blob/main/examples/vqa-ingredient.ipynb}{https://github.com/parkervg/blendsql/blob/main/examples/vqa-ingredient.ipynb}}, as shown in \citet{bae2024ehrxqa}. This section presents an example of how adding of a VQA Ingredient to BlendSQL allows us to do QA over a different unstructured media (images). In this example, a table and the corresponding image from a Wikipedia page\footnote{\href{https://en.wikipedia.org/wiki/Fountains_in_Portland,_Oregon}{https://en.wikipedia.org/wiki/Fountains\_in\_Portland,\_Oregon}} are used for answering a question. Figure \ref{app_fig:vqa_table} shows the table from the Wikipedia page containing information about different fountains along with the corresponding image of the same. A tiny VQA model\footnote{\href{https://huggingface.co/bczhou/tiny-llava-v1-hf}{https://huggingface.co/bczhou/tiny-llava-v1-hf}} is used for answering the question once the correct image has been retrieved. The byte representation of the images is stored in the SQLite database and used for the question answering task.

Below, we show an example natural language to \textsc{BlendSQL} pairing for this database context. 
The below query corresponds to the question ``How many animals are in the fountain designed by Georgia Gerber?''. When executed, it invokes the VQA model on the retrieved database content to return ``There are three animals in the fountain''.
% The query in \ref{vqa_example_code} returns the correct row containing the image, and the VQA model uses the retrieved image to answer the question.

\begin{figure}[h!]
    \centering
    \includegraphics[width=\columnwidth]{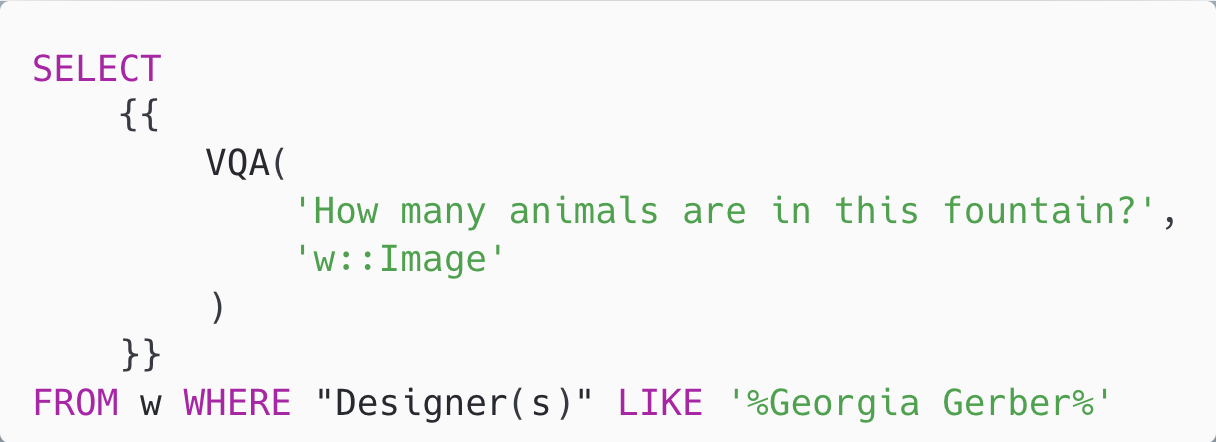}
    % \caption{\textsc{BlendSQL} for ``How many animals are in the fountain designed by Georgia Gerber?''\\The executed query returns ``There are three animals in the fountain''.}
    \label{vqa_example_code}
\end{figure}

\begin{table*}[t]
    % \centering
    \begin{tabular}{p{11em}p{29em}}
        \textbf{Q \& A} &
        \textbf{BlendSQL} \\ 
        \hline
        \begin{tabular}[c]{l}
            \small{Q: The 1995 Tooheys 1000 driver} \\ 
            \small{who was second-to-last in the} \\ 
            \small{Tooheys Top 10 was born where ?}\\ \\ 
            \small{A: Sydney}
        \end{tabular} &
        \begin{tabular}[c]{l}
\includegraphics[scale=0.2]{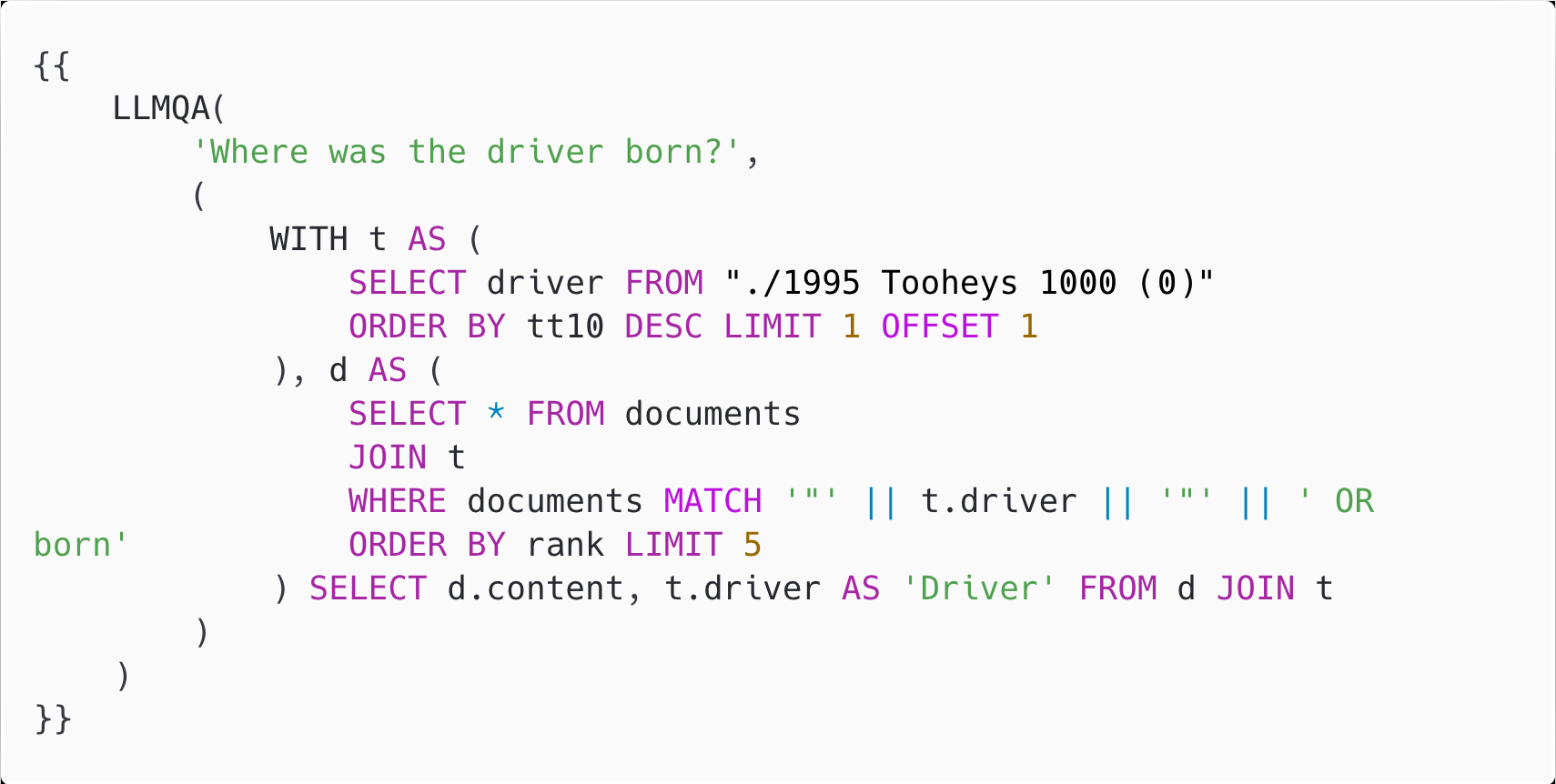}
                \\
            \small{Output: Sydney {\color{green}\checkmark}}
        \end{tabular} \\ 
        \hline
      
        \begin{tabular}[c]{l}
            \small{Q: What is the title for the} \\ 
            \small{Taiwanese television series where} \\ 
            \small{Jin Chao-chun plays a Chinese} \\ 
            \small{politician who was born in the}\\ 
            \small{year 1090 ?}\\ \\ 
            
            \small{A: Eight Thousand Li}\\
            \small{of Cloud and Moon}
        \end{tabular} &
        \begin{tabular}[c]{l}
            \includegraphics[scale=0.2]{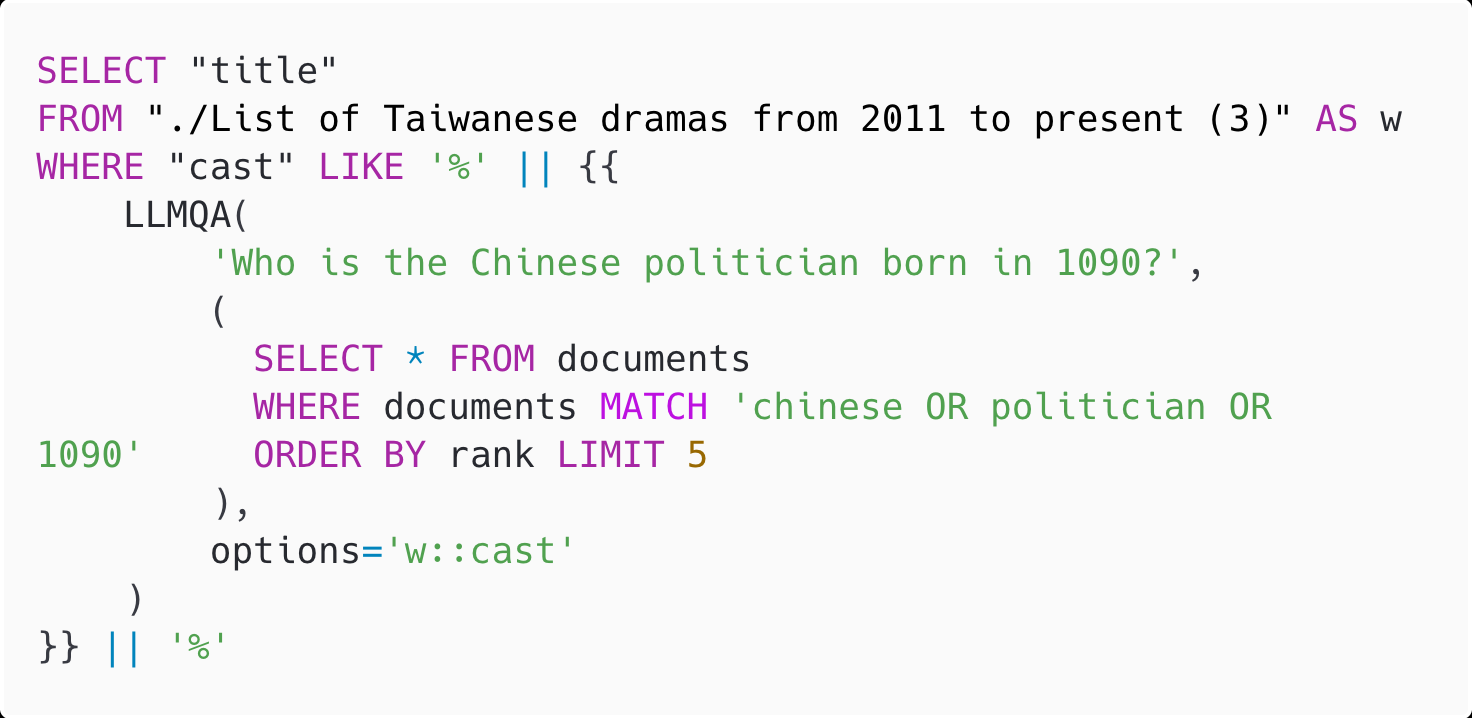}
                \\
            \small{While the logic for querying the table of unstructured content is correct,} \\ 
            \small{the model grounds the response to the wrong table. The 'role' column of} \\ 
            \small{"./Jin Chao-chun (0)" should have been used instead.} \\\\
            \small{Output: My Hero , My Daddy {\color{red}\xmark}}
        \end{tabular} \\ 
        \hline
    
        \begin{tabular}[c]{l}
            \small{Q: Abdul Hai Neamati was a} \\ 
            \small{member of a political party and was} \\ 
            \small{succeeded by Bashir Baghlani.} \\ \\ 
            \small{A: SUPPORTS}
        \end{tabular} &
        \begin{tabular}[c]{l}
                \includegraphics[scale=0.2]{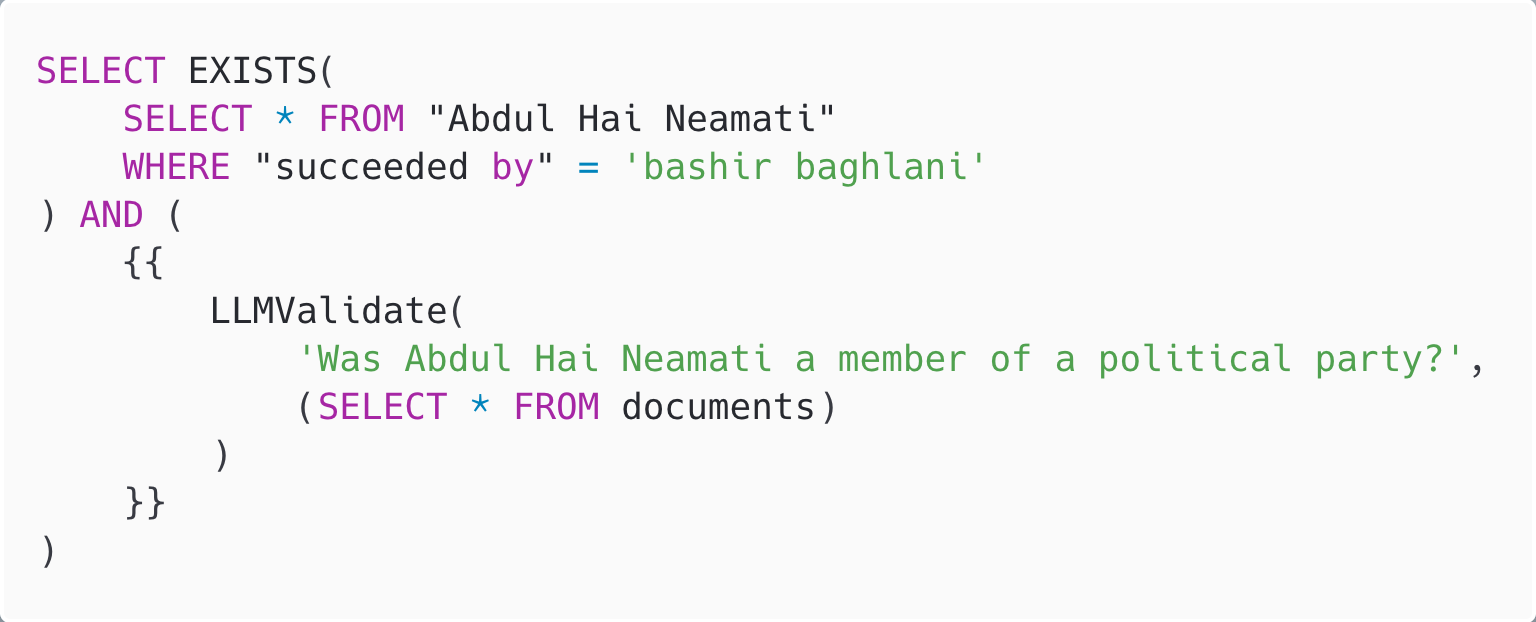}
                \\
            \small{Output: SUPPORTS {\color{green}\checkmark}}
        \end{tabular} \\ 
        \hline
    \end{tabular}
    \caption{Example outputs from OTT-QA and FEVEROUS.}
    \label{example_predictions}
\end{table*}

% \begin{table*}[t]
%     \begin{tabular}[c]{l}
      
%         \begin{lstlisting}[language=SQL,breaklines=true,showstringspaces=false,linewidth=16cm]
% SELECT w.title, w."designer ( s )", {{VQA('How many animals are in this fountain?', 'images::img_bytes')}}
% FROM images JOIN w ON w.title = images.title
% WHERE "designer ( s )" = 'georgia gerber' 
%         \end{lstlisting}\\ \\ 
%         \small{The generated SQL query returns the correct row containing the image, and the VQA model uses the retrieved image to answer} \\
%         \small{the question} \\
%         \small{Output: There are three animals in the fountain {\color{green}\checkmark}}
%     \end{tabular} \\ 
%     \caption{The BlendSQL output of the query \textit{How many animals are in the fountain designed by Georgia Gerber?}}
%     \label{app_tab:vqa_output}
% \end{table*}

\end{document}